\documentclass[pdflatex,sn-mathphys-num]{sn-jnl}

\usepackage{parskip}        
\usepackage{amsmath, amssymb}  
\usepackage{graphicx}       
\usepackage{float}          
\usepackage{caption}        
\usepackage{subfigure}      
\usepackage{subcaption}
\usepackage{rotating}       
\usepackage{pdflscape}      
\usepackage{lscape}         

\usepackage{booktabs}       
\usepackage{longtable}      
\usepackage{multirow}       
\usepackage{tabularx}       
\usepackage{tabulary}       
\usepackage{tabularray}     

\usepackage{algorithm}      
\usepackage{algpseudocode}  

\usepackage{xcolor}         
\usepackage{soul}           

\usepackage{url}            
\usepackage{hyperref}       
\hypersetup{
    colorlinks   = true,
    linkcolor    = blue,
    urlcolor     = blue,
    citecolor    = teal
}

\theoremstyle{thmstyleone}%
\theoremstyle{thmstyletwo}%

\theoremstyle{thmstylethree}%

\begin{document}

\title[Article Title]{Comparative Analysis of Deep Learning Models for Perception in Autonomous Vehicles}

\author[]{\fnm{Jalal} \sur{Khan}}\email{mjalal@uaeu.ac.ae}

\affil[]{\orgdiv{College of Information Technology}, \orgname{United Arab Emirates University, United Arab Emirates}, \orgaddress{\city{Al Ain}, \country{UAE}}}


\abstract{Recently, a plethora of machine learning (ML) and deep learning (DL) algorithms have been proposed to achieve the efficiency, safety, and reliability of autonomous vehicles (AVs). The AVs use a perception system to detect, localize, and identify other vehicles, pedestrians, and road signs to perform safe navigation and decision-making. In this paper, we compare the performance of DL models, including YOLO-NAS and YOLOv8, for a detection-based perception task. We capture a custom dataset and experiment with both DL models using our custom dataset. Our analysis reveals that the YOLOv8s model saves 75\% of training time compared to the YOLO-NAS model. In addition, the YOLOv8s model (83\%) outperforms the YOLO-NAS model (81\%) when the target is to achieve the highest object detection accuracy. These comparative analyses of these new emerging DL models will allow the relevant research community to understand the models' performance under real-world use case scenarios.}

\keywords{Deep learning; YOLO-NAS; YOLOv8; Perception; Autonomous vehicles.}


\maketitle

\section{Introduction}
\label{sec:Intro}
The United States (DARPA Agency) and Europe (EUREKA Prometheus Project) made significant contributions to enhance the necessary technologies, research, and development for autonomous driving (AD) technology~\cite{DARPAUrb5:online, EurekaPr8:online,khan2022level}. In addition, the success of deep neural networks in ImageNet allowed relevant research communities to propose significant deep learning (DL) models to detect objects for the perception system of autonomous vehicles (AVs)~\cite{khan2023augmenting,mao20233d,malik2023should}. For instance, Sparse4D-v3, Far3D, HoP, Li, StreamPETR-Large, VCD-A, UniM2AE, SparseBEV, and VideoBEV techniques are proposed for object detection based on nuScenes dataset during the year 2023~\cite{Objectde0:online}. These models introduce various unique approaches to perception tasks and are designed to improve accuracy and speed in challenging environments. For object detection, the newly available DL-based architectures, i.e., the YOLOv8 and YOLO-NAS models, are compared on the COCO dataset using NVIDIA T4~\cite{GitHubDe57:online}. However, these models have not yet been extensively compared and evaluated for performance analysis using a custom road dataset. In this study, we investigate the impact of DL models on object-detection-based perception tasks for AVs. In this connection, we consider a custom road dataset from \cite{khan2024vehicle}. We compared and evaluated YOLOv8s and YOLO-NAS using well-known performance metrics, such as mean average precision (mAP), and confusion matrix values. The contributions of this research work are comparing the performance of DL models used by AVs for object detection tasks.

\section{Methodology}
\label{sec:methodology}
The perception system of AVs perceives road objects using state-of-the-art (SOTA) DL-based models. Since the YOLOv8s and YOLO-NAS-based DL models have not been explored extensively, we select them for experimental work considering object detection-based perception tasks of AVs. For object detection tasks, we use a camera as the primary capturing device to capture data (video clips) on complex urban road segments. We record a total of 15,100 images during the data-capturing phase. Next, we reduce the initial capture dataset to 1000 images using different pre-processing techniques. We manually label our final dataset, which contains five different classes i.e., cars, pedestrians, motorcyclists, trucks, and rickshaws. In addition, we split the dataset into a train set (70\%), a validation set (20\%), and a test set (10\%). The rationale behind the YOLOv8s and YOLO-NAS is that the selection of a single DL model depends on the underlying use case scenarios~\cite{mao20233d}. Now that we have our pre-processed dataset and DL-based models, we start to perform the training, validation, and testing of our models. It should be highlighted that we use the basic hyperparameters (i.e., 100 epochs, 16 batch size, etc.) to get a fair performance comparison of both DL models. Now that we have our custom-trained DL-based models, we start investigating the performance of each model.

\section{Experiments}
\label{sec:experiments}
The experiments were conducted using the NVIDIA Tesla V100 GPU model, 16.0GB of GPU RAM, 12.7GB of System RAM, etc., using Google Colab Pro platform. The NVIDIA Tesla V100 GPU was accessed through Google Chrome (version 117.0) installed in a physical machine of Intel Quad-Core i7@ 3.4GHz with 32GB NVIDIA GeForce GTX 680MX Graphics and a 27-inch (2560x1440) build-in Display. The software stack includes Python v.3.11.4 (the newest major release), and its rich family of libraries e.g., PyTorch, glob, etc. The YOLOv8s and YOLO-NAS based DL models were implemented through their respective packages using the latest version of Python programming language. The learning model instances were trained on image size of 640x640. For evaluation, we rely on well-known mean average precision (mAP) metric, which provide answer to the following question: How consistently accurate are the learning model's predictions across different recall levels and across all classes?

\begin{figure}[t]
\centering
\includegraphics[width=\columnwidth]{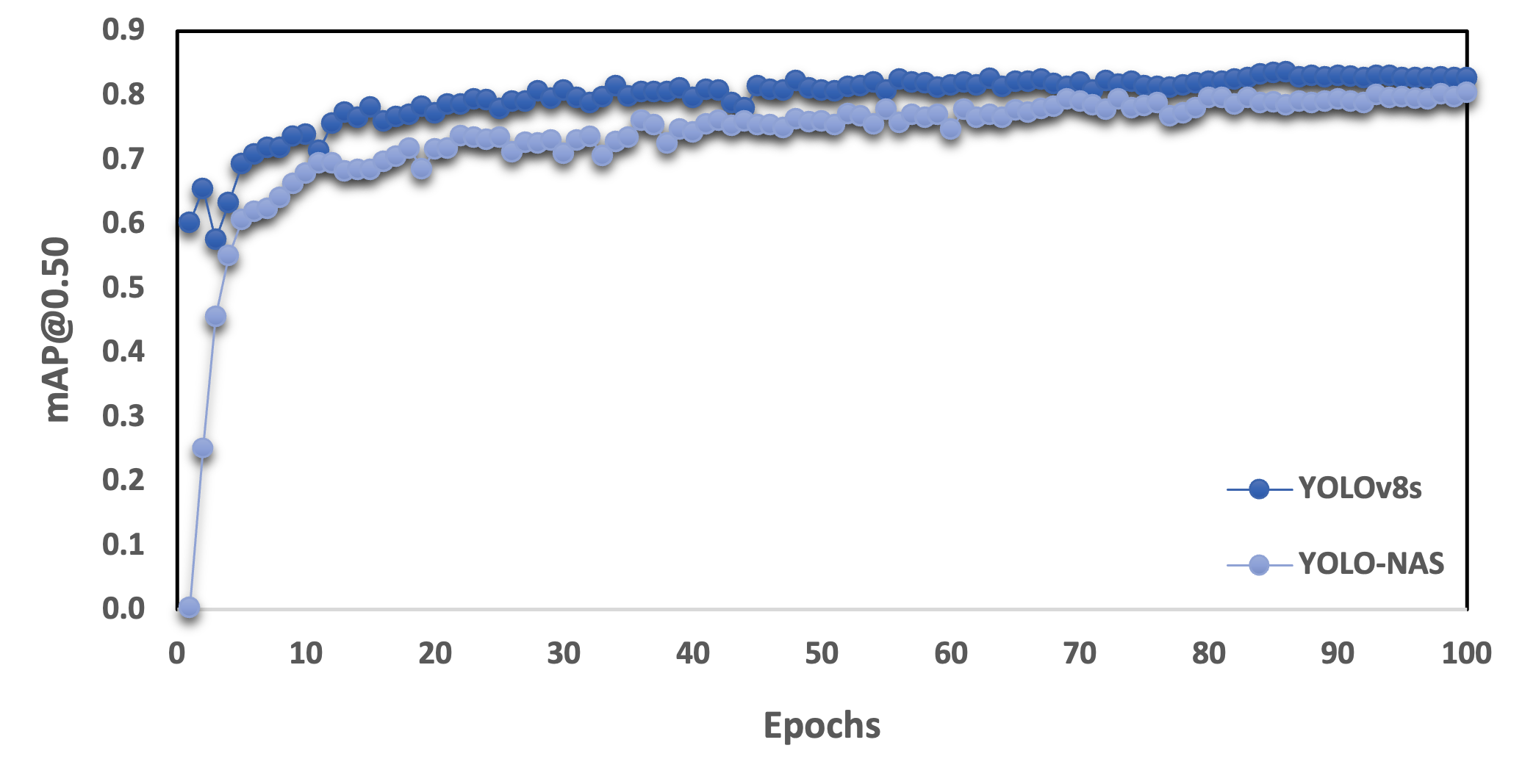} 
\caption{Comparison of mAP@0.5 Performance Metrics.}
\label{mAPs}
\end{figure}

\section{Performance Evaluation}
\label{sec:PE}
The purpose of conducting these experiments is to investigate the behavior and performance of DL models custom-trained over GPU-capable devices. We evaluated the performance of object detection task for the perception system of the AVs. The uniform setting and configuration are used for the experiments with both DL-based models. All experiments are conducted on the same custom dataset (as discussed in Section~\ref{sec:methodology}) and using the custom-trained YOLOv8s and YOLO-NAS DL models with configuration i.e., 16 batch size, 100 epochs. In what follows next, we discuss the findings concluded from the experimental results.

\begin{figure}[]
\centering
\includegraphics[width=8cm]{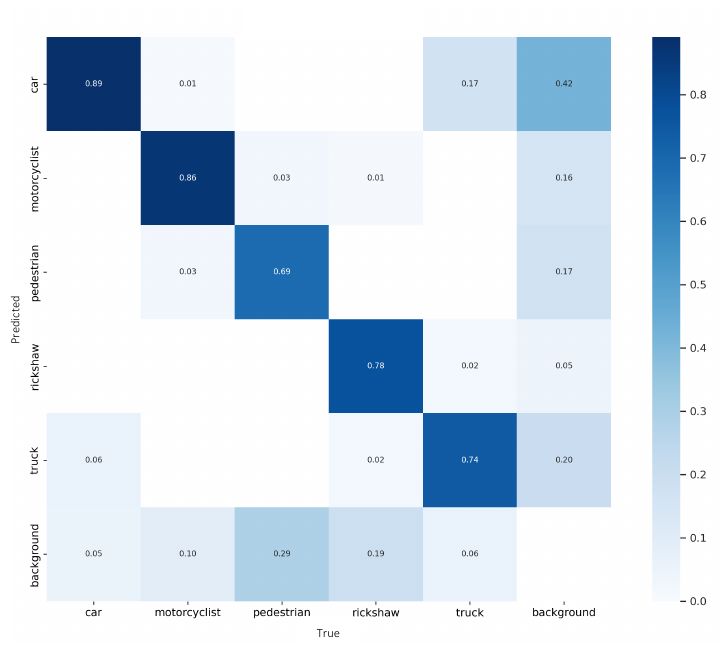} 
\caption{Confusion Matrix for YOLOv8s Model.}
\label{y8v}
\end{figure}

\begin{figure}[]
\centering
\includegraphics[width=8cm]{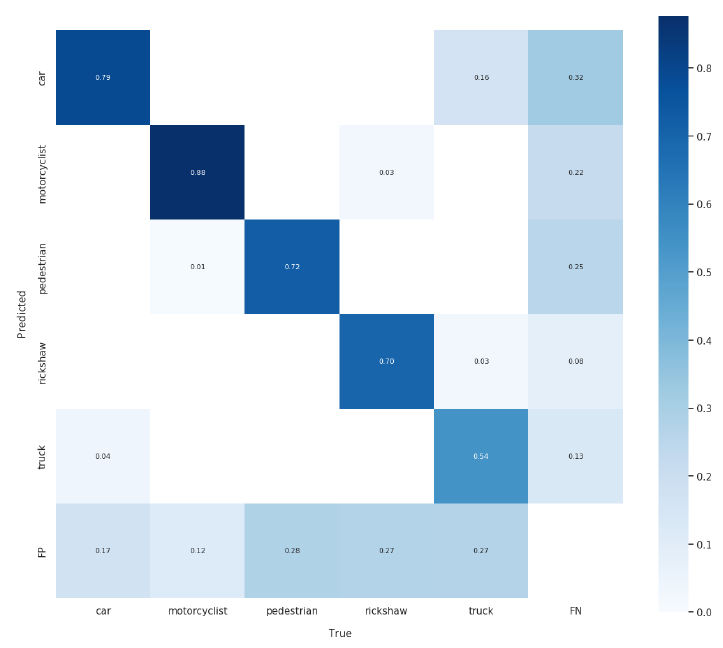} 
\caption{Confusion Matrix for YOLO-NAS Model.}
\label{y8-nas}
\end{figure}

The performance of both the DL models for a detection-based perception task is shown in Figure~\ref{mAPs}. The YOLOv8s model starts with a high mAP of 0.602, whereas, the YOLO-NAS model starts with a low mAP of 0.003. Meaning thereby, the YOLOv8s outperforms YOLO-NAS for initial performance until their fifth epoch. Furthermore, the learning curve of the YOLOv8s model shows a steady increase with consistent improvement and fewer fluctuations. Whereas, the learning curve of the YOLO-NAS model shows a sudden improvement in the beginning and less consistency in comparison to the YOLOv8s model. It is important to highlight that the YOLOv8s model shows a stable and reliable performance in comparison to the YOLO-NAS model. Therefore, the YOLOv8s model is more robust and effective for object detection-based perception task. 

To evaluate the performance further, we rely on the confusion matrices of both DL models as shown in Figure~\ref{y8v} and Figure~\ref{y8-nas}. For the car objects, the YOLOv8s predicted cars 89\% of the time and the YOLO-NAS predicted cars 79\% of the time. For the car objects, the true predictions are 89\% by the YOLOv8s model and 79\% by the YOLO-NAS model. It is also evident from the false positives (background), where the car object has 0.05 by the YOLOv8s model and 0.17 by the YOLO-NAS model. In a nutshell, the YOLOv8s model is a suitable choice for car, rickshaws, and trucks. Similarly, the YOLO-NAS model is a good choice for motorcyclists and pedestrian objects. To achieve safe navigation by the AVs, both the DL models can be integrated for object detection tasks performed by the perception system.

\section{Conclusions}
\label{sec:conc}
The selection of a DL technique is imperative to improve safe navigation and efficiency of the perception system used in AVs. In this paper, we experimentally investigate the SOTA DL-based YOLOv8s and YOLO-NAS models for the perception system of AVs. For comparisons, we applied performance metrics: mAPs, and confusion matrix. Consequently, the YOLOv8s model outperforms the YOLO-NAS model in terms of stability, reliability, and accurate performance.

\section*{Acknowledgment}
The author would like to acknowledge the United Arab Emirates University for providing the necessary resources.

\end{document}